\documentclass{article} 
\usepackage[preprint]{neurips_2024}

\usepackage{amsmath,amsfonts,bm}

\def\eqref#1{equation~\ref{#1}}

\def\1{\bm{1}}

\DeclareMathAlphabet{\mathsfit}{\encodingdefault}{\sfdefault}{m}{sl}
\SetMathAlphabet{\mathsfit}{bold}{\encodingdefault}{\sfdefault}{bx}{n}

\usepackage{microtype}
\usepackage{pifont}

\title{
WildLong: Synthesizing Realistic \\
Long-Context Instruction Data at Scale
}

\author{
Jiaxi Li$^{\dagger}$\thanks{Work done during internship at Microsoft Research.} \quad Xingxing Zhang$^\ddagger$ \quad Xun Wang$^\ddagger$ \quad Xiaolong Huang$^\ddagger$ \quad Li Dong$^\ddagger$ \quad Liang Wang$^\ddagger$\\
\textbf{Si-Qing Chen}$^\ddagger$ \quad \textbf{Wei Lu}$^\dagger$ \quad \textbf{Furu Wei}$^\ddagger$
\\$^\dagger$Singapore University of Technology and Design \quad $^\ddagger$Microsoft Research\\
}

\usepackage[hidelinks]{hyperref}
\usepackage{url}
\usepackage{graphicx}
\usepackage{paralist}
\usepackage{url}
\usepackage{xspace}
\usepackage{amsfonts,amsmath}
\usepackage{booktabs}
\usepackage{setspace}
\usepackage{multirow}
\usepackage{multicol}
\usepackage{subfigure}
\usepackage{wrapfig}
\usepackage{lipsum}
\usepackage{booktabs}
\usepackage{makecell}
\usepackage{amssymb}
\usepackage{pifont}
\newcommand{\cmark}{\ding{52}}%
\newcommand{\ostar}{\ding{72}}
\usepackage{color, colortbl, xcolor}
\usepackage{dashrule}
\usepackage[most]{tcolorbox}

\definecolor{mydarkblue}{rgb}{0,0.08,0.45}
\definecolor{mydarkgreen}{HTML}{32612D}
\definecolor{myblue}{HTML}{3b75c3}
\definecolor{myred}{HTML}{E33222}
\definecolor{mygreen}{HTML}{438773}
\definecolor{mymaroon}{RGB}{142,27,19}
\definecolor{maroon}{HTML}{800000}
\definecolor{mycite}{cmyk}{0.55,1,0,0.15}
\definecolor{codeblue}{rgb}{0.25,0.5,0.5}
\definecolor{codekw}{rgb}{0.85, 0.18, 0.50}
\definecolor{codegreen}{rgb}{0,0.6,0}
\definecolor{codegray}{rgb}{0.5,0.5,0.5}
\definecolor{codepurple}{rgb}{0.58,0,0.82}
\definecolor{backcolour}{rgb}{0.95,0.95,0.92}
\hypersetup{
    colorlinks=true,
    citecolor=mydarkblue,
    linkcolor=codepurple,
    urlcolor=maroon
          }

\definecolor{gr}{RGB}{0, 146, 0}

\usepackage{bbm}
\usepackage{wrapfig}
\usepackage{lipsum}
\usepackage{enumitem}
\usepackage{paralist, tabularx}

\begin{document}
\maketitle

\begin{abstract}
Large language models (LLMs) with extended context windows enable tasks requiring extensive information integration but are limited by the scarcity of high-quality, diverse datasets for long-context instruction tuning. Existing data synthesis methods focus narrowly on objectives like fact retrieval and summarization, restricting their generalizability to complex, real-world tasks.
WildLong extracts meta-information from real user queries, models co-occurrence relationships via graph-based methods, and employs adaptive generation to produce scalable data. It extends beyond single-document tasks to support multi-document reasoning, such as cross-document comparison and aggregation.
Our models, finetuned on 150K instruction-response pairs synthesized using WildLong, surpasses existing open-source long-context-optimized models across benchmarks while maintaining strong performance on short-context tasks without incorporating supplementary short-context data.
By generating a more diverse and realistic long-context instruction dataset, WildLong enhances LLMs' ability to generalize to complex, real-world reasoning over long contexts, establishing a new paradigm for long-context data synthesis.
\end{abstract}

\section{Introduction}

The growing demand for AI systems capable of processing and reasoning over extensive information has driven the development of large language models (LLMs) with significantly expanded context windows \citep{dubey2024llama, achiam2023gpt, team2024gemini}. 
Among long-context tasks, needle-in-a-haystack (NIAH) \citep{niah} retrieval—where models locate specific information within large contexts—has emerged as a relatively simple benchmark, with previous work showing that continued pretraining on long-context data significantly boosts NIAH performance \citep{fudata, hsieh2024ruler, li2024needlebenchllmsretrievalreasoning}.
However, while many LLMs excel at NIAH, they often struggle with more complex challenges, such as passage ranking and dialogue analysis, which require reasoning and synthesis across extended contexts \citep{hsieh2024ruler, yen2025helmet, zhang2024infinitebench, levy-etal-2024-task, vodrahalli2024michelangelolongcontextevaluations, li2024alr2retrievethenreasonframeworklongcontext}. 
The ability to reason over long contexts is essential for real-world applications, such as legal document analysis and book review \citep{liu2024lost, karpinska2024one, xuretrieval, xu2024detectiveqa, jimenez2024swebench, wang-etal-2024-leave}.
These more difficult tasks require models not only to retrieve information but also to integrate and reason over it in realistic, multi-faceted scenarios. Addressing this gap calls for high-quality, diverse, and generalized instruction-tuning datasets designed specifically for long-context reasoning. Such datasets are essential for equipping LLMs to effectively leverage their extended context capabilities in complex, real-world applications.

A major bottleneck in enhancing long-context reasoning is the lack of high-quality instruction tuning data. Unlike short-context tuning, which benefits from abundant human-annotated data, manually constructing long-context instruction data is impractical due to the complexity of reasoning over extended contexts.
Existing methods rely on data synthesis using LLMs \citep{dubey2024llama, an2024make, bai2024longalign, xiong2024effective, xiong2024artificialneedlesrealhaystacks}. 
For instance, prior approaches \citep{xiong2024effective, bai2024longalign} generate long-context instruction-tuning data by extracting short text spans from long documents, synthesizing question-answer pairs based on these snippets, and incorporating the full document during training. Other approaches, such as Llama-3.1 \citep{dubey2024llama}, further utilize hierarchical summarization to construct long-context datasets.
While effective at leveraging models' short-context capabilities for data generation, these methods primarily focus on fact extraction and summarization.
This narrow scope limits the diversity and generalizability of the resulting data, leaving critical gaps in supporting more complex and realistic tasks.

\begin{figure*}[t]
    \centering
    \includegraphics[width=0.99\textwidth]{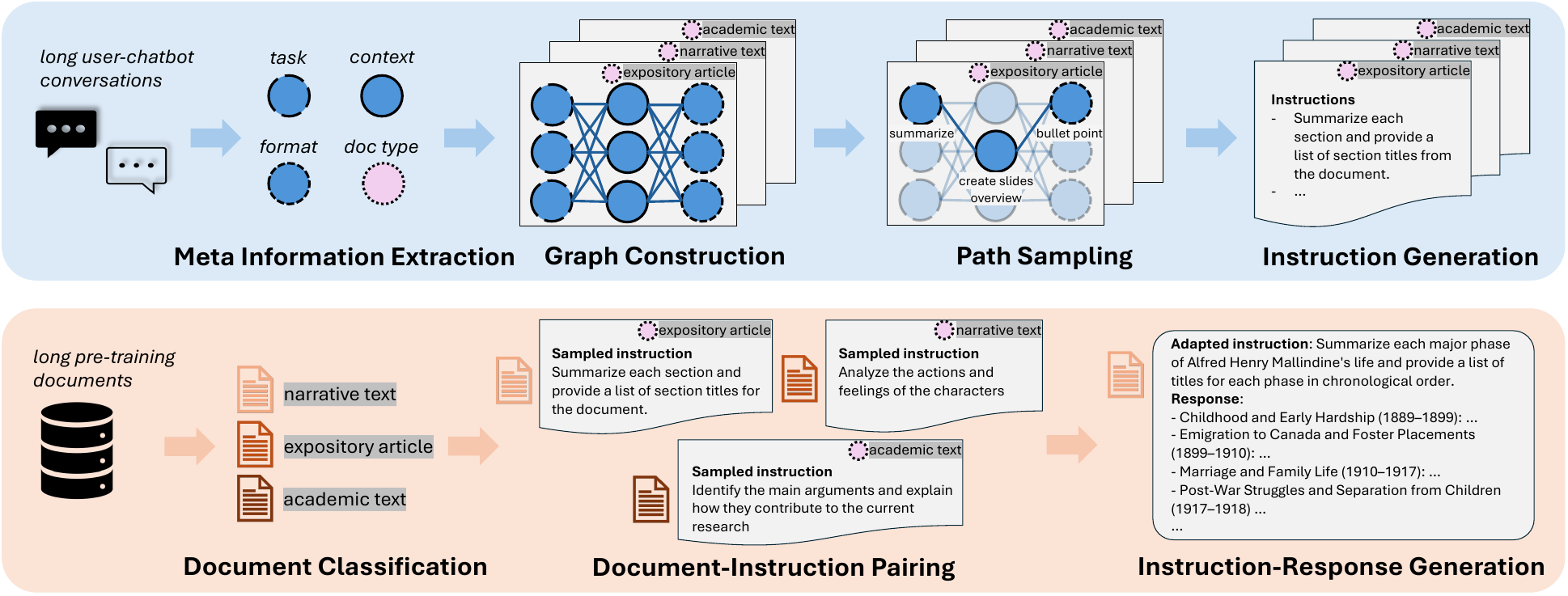}
    \caption{Overview of the two-stage WildLong Framework. Stage 1 extracts meta-information from real-world user-chatbot conversations, classifies documents by type, constructs graphs to represent meta-information relationships, and samples paths to generate tailored instructions. Stage 2 pairs long documents from the pre-training corpus with these instructions, generating instruction-response pairs by rewriting the instructions and answering based on the document context.}
    \label{fig:method_overview}
\end{figure*}

To address this limitation, we propose WildLong, a scalable framework for generating diverse and realistic instruction-response pairs for long-context reasoning. Our approach integrates \textit{meta-information extraction}, \textit{graph-based modeling}, and \textit{adaptive instruction-response generation}. The pipeline of our framework is illustrated in Figure~\ref{fig:method_overview}.
We extract meta-information, such as user intents, tasks, and constraints, from real-world user-chatbot conversations. This process ensures that the generated instruction-response pairs are grounded in realistic interactions and reflect the complexity of real-world scenarios.
To enhance diversity and scalability, we model the extracted meta-information as a graph, where nodes represent individual meta-information value and edges capture their co-occurrence frequencies. By performing random walks on this graph, we generate novel combinations of meta-information, introducing diverse and varied instruction templates. 
Adaptive instruction-response generation further supports scalability and diversity. Each combination of meta-information is paired with long-context examples sampled from the pretraining corpus, introducing variability in the contexts associated with the instructions. The availability of abundant pretraining data ensures that large-scale datasets can be generated efficiently.
As shown in Figure \ref{fig:task_doc_type_distribution}, our dataset spans a wide range of document types and task types, reflecting the diversity and complexity required for real-world long-context reasoning.

We fine-tuned Mistral-7B-Instruct-v0.2\footnote{\url{https://huggingface.co/mistralai/Mistral-7B-Instruct-v0.2}} and Llama-3.1-8B-Instruct\footnote{\url{https://huggingface.co/meta-llama/Llama-3.1-8B-Instruct}} on 150K synthesized instruction-response pairs and evaluated them on various long-context benchmarks with input lengths up to 128K tokens.
Notably, our fine-tuned Mistral-7B model achieves a substantial $+14.7$ improvement on the RULER benchmark \citep{hsieh2024ruler}, while our Llama-3.1-8B model performed competitively with much larger models, scoring $84.1$ on RULER (vs. $85.1$ for Llama-3.1 70B) and $6.8$ on LongBench-Chat \citep{bai2024longalign} (vs. $6.7$ for Llama-3.1 70B). 
Importantly, our fine-tuned models retain short-context performance without fine-tuning on additional short-context data, which existing methods typically use to prevent degradation. This demonstrates the robustness and generalizability of our synthetic data.



\section{Proposed Method}

\begin{figure*}[t]
    \centering
    \includegraphics[width=0.99\textwidth]{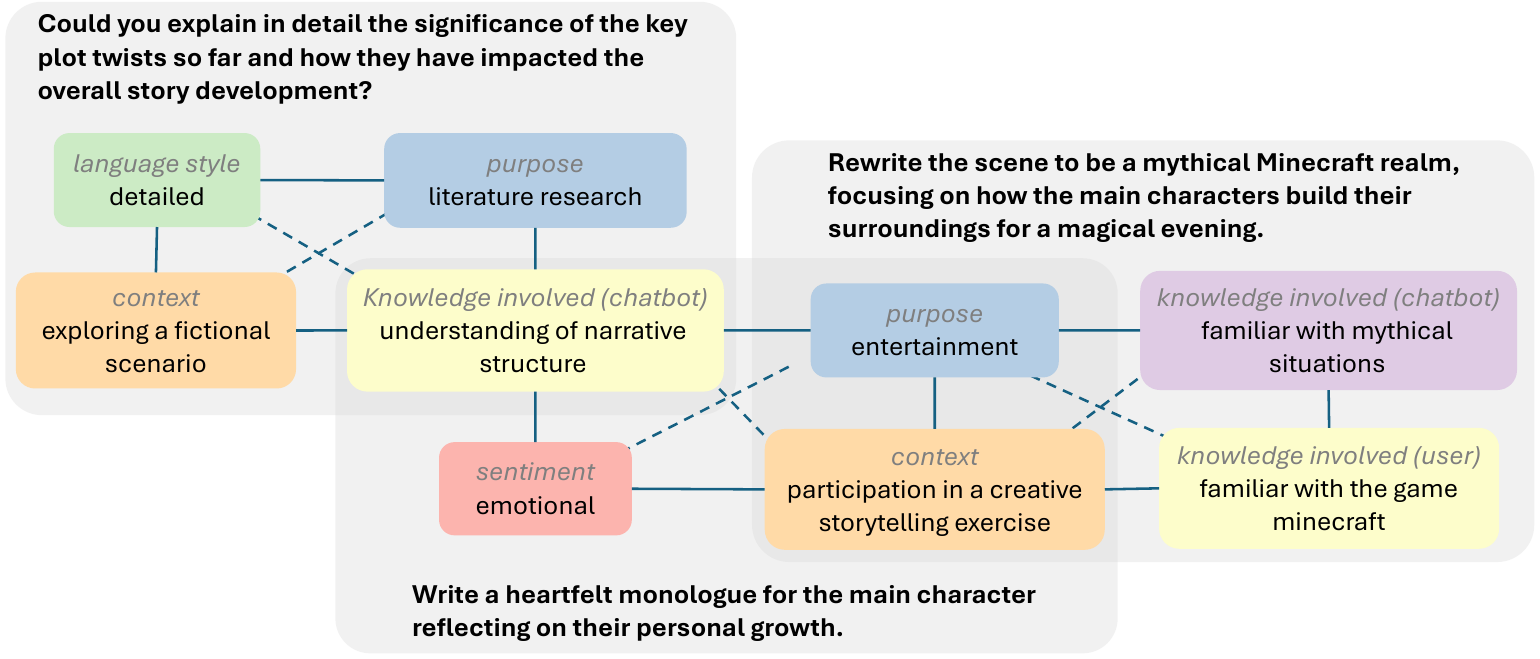}
    \caption{This figure demonstrates examples of instructions generated from sampled paths in a narrative text graph. Solid lines represent connections within paths, while dotted lines show node interconnections during graph construction. Using a random walk algorithm, diverse instructions are generated by combining nodes. For instance, the knowledge node ``understanding of narrative structure'' and the context node ``participation in a creative storytelling exercise'' appear in multiple paths but result in distinct instructions due to varying other meta information.}
    \label{fig:path_example}
\end{figure*}

In this section, we describe our methodology for generating diverse and realistic instruction-response pairs for long-context tasks. 
As shown in Figure \ref{fig:method_overview}, our approach comprises two main stages. In Stage 1, we extract meta-information from real-world user-chatbot conversations and construct document-type-specific graphs to model co-occurrences among meta-information. Instructions are generated by sampling paths from these graphs. In Stage 2, these instructions are paired with long documents from the pretraining corpus to create instruction-response pairs. Below, we provide an overview of each component in the framework.

\subsection{Meta Information Extraction}
We leverage the WildChat dataset \citep{zhao2024wildchat}, a large corpus of user-chatbot conversations, and focus specifically on single-turn conversations that involve long contexts. WildChat is particularly suitable for our task because it contains realistic user queries and high-quality responses, which facilitate the accurate annotation of meta information fields.
From each conversation, we extract 13 fields of meta information that represent key attributes relevant to understanding and modeling long-context instructions:

\small{\texttt{document type, tasks or requests, user intention, user profile, language style, context, knowledge/commonsense involved for user, knowledge/commonsense involved for chatbot, long context capability involved, output format, sentiment, constraint of the request, simplified instruction}}.

These fields encompass essential aspects of the interaction, ensuring a comprehensive representation of user intent, contextual nuances, and task-specific requirements.
We prompt {\tt GPT-4} to extract meta information from each conversation\footnote{The prompt used for meta information extraction is detailed in Table \ref{tab:prompt_extract_meta}.}. For example, tasks such as ``extract details'' for informational articles or ``continue the story'' for fictional narratives are explicitly labeled. Contexts like ``preparing for a presentation'' or ``research related to ancient Greece'' are extracted for professional or historical texts, respectively.
The extracted meta information reflects realistic user scenarios involving long-context conversations and serves as a structured foundation for subsequent stages of our methodology.


\subsection{Graph Construction}
\begin{wrapfigure}{r}{0.5\textwidth}
    \vspace{-0.2in}
    \centering
    \includegraphics[width=0.9\linewidth]{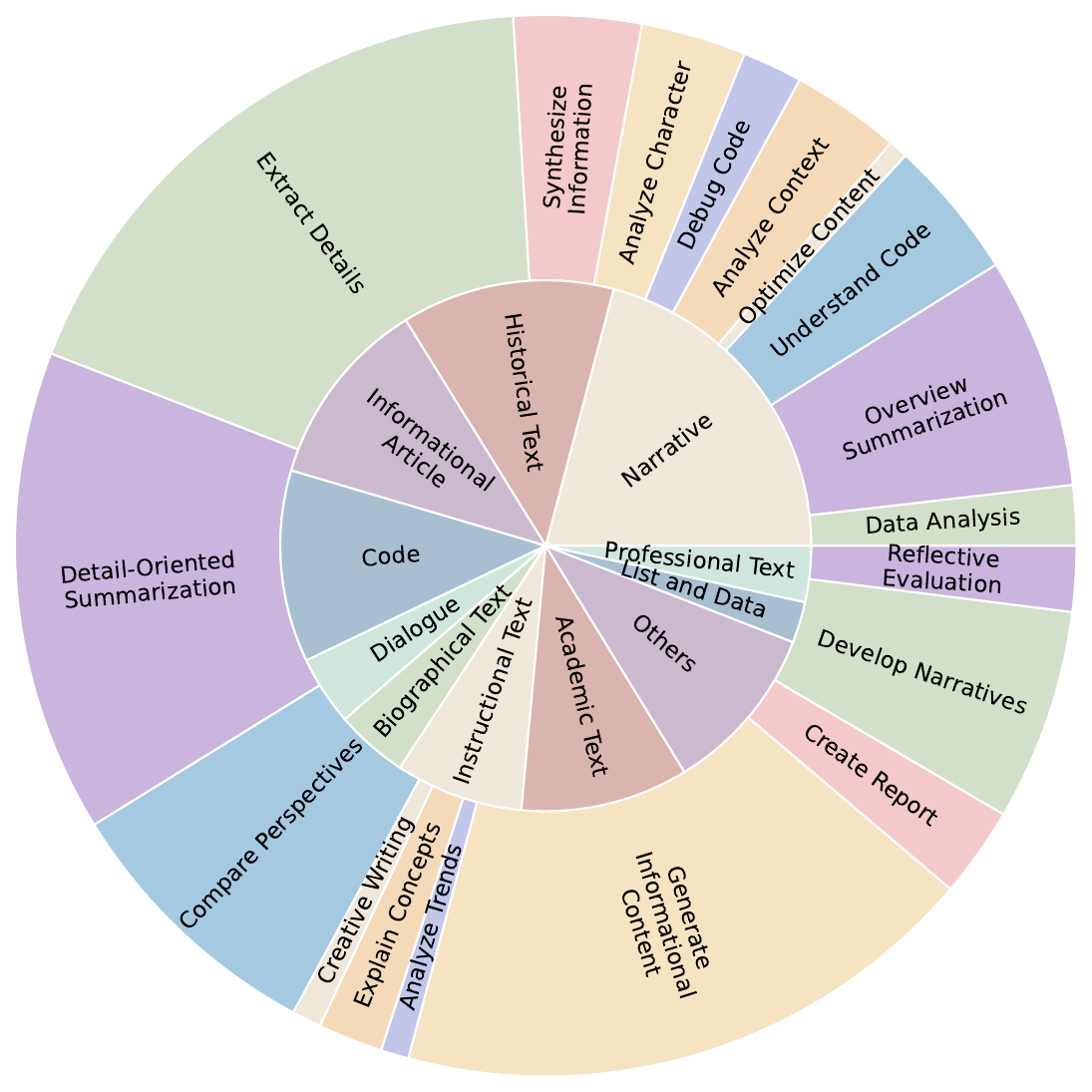}
    \caption{Distribution of document types (inner circle) and task types (outer circle) in our dataset.}
    \label{fig:task_doc_type_distribution}
    \vspace{-0.2in}
\end{wrapfigure}

Instructions are generally document-type-specific, necessitating the construction of separate graphs for each document type. 
To build document-type-specific graphs, we first identify document types for each conversation as free-form values during the meta information extraction process. 
To group these values into coherent and meaningful categories, we apply K-Means clustering. The total number of clusters, set to 10, is predefined to balance between generalization and specificity based on the observed diversity of the dataset. Each cluster represents a distinct document type, and the cluster centers are rewritten to serve as the final document type labels. The distribution of the document types is illustrated in Figure \ref{fig:task_doc_type_distribution}. 

For each document type $d$, we construct an undirected graph $G_d = (\mathbb{V}_d, E_d)$ to model the co-occurrence relationships among meta information values extracted from user-chatbot conversations\footnote{Eleven meta information fields are used to construct the graph. The ``document type'' field is used to classify documents such that we can construct a separate graph for each document type. The ``simplified instruction'' field is used as a demonstration when generating instructions based on paths, see Section \ref{sec:path_to_instructs}.}. This graph represents the interactions between meta information fields and facilitates the systematic exploration of realistic and diverse combinations for instruction generation. The construction process is detailed as follows.

\paragraph{Nodes}
Each node corresponds to a unique value of a meta information field. Let $\mathbb{M} = \{m_1, m_2, \dots, m_{11}\}$ denote the set of 11 meta information fields used to construct the graph (e.g., task type, sentiment, output format). The set of nodes $\mathbb{V}_d$ is defined as:
\[
\mathbb{V}_d = \{v \mid v \text{ is a value of some field } m_i \in \mathbb{M}\text{ in any conversation for document type } d\}.
\]
Nodes are independent of individual conversations and collectively capture all unique meta information values observed for the document type.

\paragraph{Edges}
Edges represent the co-occurrence of meta information values in the same conversation, provided they belong to different fields.
Formally, an edge $(v, u) \in E_d$ exists if:
\begin{enumerate}
    \item $v$ is a value of field $m_i \in \mathbb{M}$,
    \item $u$ is a value of field $m_j \in \mathbb{M}$, where $i \neq j$, and  
    \item $v$ and $u$ co-occur in at least one conversation for document type $d$.
\end{enumerate}
For each conversation, the extracted meta information values from the 11 fields are interconnected, forming a fully connected bipartite subgraph, where edges connect values from different fields.

\paragraph{Edge Weights}
The weight of an edge $(v, u) \in E_d$ reflects the frequency of co-occurrence of $v$ and $u$ across all conversations for document type $d$. The edge weight is computed as:
\[
w(v, u) = \log(f_{\text{co}}(v, u) + \varepsilon),
\]
where $f_{\text{co}}(v, u)$ is the raw count of co-occurrences, and $\varepsilon$ is a small constant for numerical stability. The logarithmic scaling mitigates the influence of highly frequent pairs while preserving distinctions among lower-frequency edges.

We build document-type-specific graphs by fully connecting meta information values co-occurring in the same conversation, with edges weighted by log-scaled co-occurrence. 
By preserving the variety of meta information and accurately capturing their co-occurrence patterns, the graph facilitates the generation of realistic, meaningful and diverse instruction paths.

\subsection{Meta Information Path Sampling}
To ensure that instruction generation is guided by realistic and diverse criteria, we first sample structured combinations of meta information values. Since meta information fields interact in complex ways, manually enumerating all meaningful combinations is infeasible. Instead, we apply a weighted random walk on the document-type-specific graphs to systematically explore plausible meta information combinations.
To generate sampled paths $\hat{P} = \{ v_1, v_2, \dots, v_k \}$ that represent meta information combinations, we employ a weighted random walk algorithm on $G_d$. 

The walk begins by randomly selecting an initial node $v_1 \in \mathbb{V}_d$ from a uniformly sampled meta information category $m_c \in \mathbb{M}$. 
At each step $t$, the walk transitions from the current node $v_t$ to a neighboring node $v_{t+1}$, which belongs to a different meta information category that has not yet been visited. The transition probability from $v_t$ to $v_{t+1}$ is determined by edge weights:
\begin{equation}
    p(v_{t+1} \mid v_t) = \frac{\exp(w(v_t, v_{t+1}))}{\sum_{v_k \in \mathcal{N}(v_t)} \exp(w(v_t, v_k))},
\end{equation}
where $w(v_t, v_{t+1})$ is the weight of the edge between $v_t$ and $v_{t+1}$, and $\mathcal{N}(v_t)$ is the set of neighbors of $v_t$. 
The walk continues for up to $N$ steps, producing a path that spans $N$ distinct meta information fields. 
Based on our preliminary experiments on instruction synthesis, we determined that $N = 6$ strikes the right balance: larger values of $N$ introduce overly restrictive criteria, making instruction generation challenging and prone to producing convoluted instructions joined by ``and'' to satisfy all requirements. Conversely, smaller values of $N$ result in overly simple instructions with limited complexity. 
The limit of six meta information fields provides sufficient criteria to guide instruction generation while allowing the model to flexibly incorporate other relevant meta information creatively.
By leveraging edge weights to guide transitions, the algorithm captures realistic co-occurrence patterns, enabling the scalable synthesis of diverse instruction templates, while maintaining flexibility to explore less frequent connections. These paths serve as structured templates to generate diverse and representative instructions for long-context tasks.

\subsection{Instruction Generation with Sampled Paths}
\label{sec:path_to_instructs}

To synthesize instructions aligned with the sampled meta-information paths, we prompt {\tt GPT-4} with a one-shot demonstration.
{\tt GPT-4} generates natural language instructions that follow the criteria defined by the meta information fields in the sampled path\footnote{Details about how to select the demonstration can be found in Appendix \ref{sec:appendix_demo} and the prompt can be found in Table \ref{tab:prompt_path_to_instruct} in Appendix \ref{sec:appendix_prompt}.}.  
Figure \ref{fig:path_example} illustrates instructions generated from sampled paths in a narrative text graph. Using the random walk algorithm, diverse instructions emerge by combining different meta information values. For example, two paths may share ``understanding of narrative structure'' as the ``knowledge involved for chatbot'' field but differ in others. One path, with values like ``detailed language style'' and ``literature research purpose,'' guides an instruction for analyzing plot development. Another, with ``entertainment purpose'' and ``emotional sentiment,'' leads to an instruction for crafting a heartfelt monologue.

\subsection{Instruction-Response Pair Generation}

Once the instructions are generated, we pair them with long documents sampled from the SlimPajama\footnote{SlimPajama is an open-source reproduction of the LLaMA pretraining data mixture \citep{touvron2023llama}.} dataset \citep{cerebras2023slimpajama}.
 SlimPajama’s wealth of long documents makes it well-suited for tasks requiring extensive context.
As instructions are document-type-specific, we first classify sampled documents into one of ten predefined document types using a custom classifier\footnote{Details about the classifier are provided in Appendix \ref{sec:appendix_doc_classifier}.}.


To ensure the paired documents reflect realistic user needs for long-context capabilities, we resample SlimPajama’s long documents to align their document type distribution with that of WildChat long conversations. This adjustment ensures the data distribution is representative of how users typically query about long contexts.
Once the document types are predicted, we pair each document with an instruction generated from the graph corresponding to its type. 
To make the instructions more contextually grounded, the sampled instruction and paired document are provided as input to {\tt GPT-4}, which generates an adapted instruction aligned with the document, and a corresponding response\footnote{Details about the prompt can be found in Table \ref{tab:prompt_instruct_response} in Appendix \ref{sec:appendix_prompt}.}. This ensures the final instruction-response pairs are coherent, relevant, and reflective of the document’s context.


\subsection{Extending Instructions to Multi-Document Settings}
We observe that the filtered WildChat dataset predominantly contains instructions designed only for single-document contexts, with limited coverage of multi-document tasks. To address this gap, we extend our method to generate instructions suitable for multi-document settings by adapting the extracted meta information and graph-based framework.

The extension begins with adapting the ``tasks or requests'' field in the meta information to reflect multi-document requirements while keeping other fields unchanged. 
Each single-document task node is rewritten to explicitly involve handling information across multiple documents using {\tt GPT-4}.
For instance, a task like ``Summarize the key points of the document'' is transformed into ``Summarize and compare the key points across multiple documents.''\footnote{The rewriting prompt is shown in Table \ref{tab:prompt_single_to_multi}. The modifications emphasize the need for synthesis, comparison, or aggregation across documents while preserving coherence and relevance. }

Following this adaptation, we construct document-type-specific graphs for multi-document tasks, sample paths using the same random walk algorithm, and generate instructions based on the sampled paths. The steps for graph construction, path sampling, and instruction synthesis remain largely consistent with the single-document setting.
During the document-instruction pairing stage, we sample pairs of documents of the same type from the SlimPajama dataset, concatenate them, and pair the concatenated documents with a multi-document instruction of the same type.
The concatenated documents and their paired instruction are then input to {\tt GPT-4} to generate a refined, contextually aligned instruction and a corresponding response.
By integrating these modifications, our method systematically generates instructions and responses that support multi-document reasoning tasks.

\section{Experiments}
We evaluate our framework comprehensively on both long-context and short-context benchmarks. This section outlines implementation details, compares our method with baseline pretrained and specialized long-context optimized models, benchmarks against existing long-context supervised fine-tuning (SFT) datasets, and presents ablation studies to analyze the contributions of essential components in our framework.

\subsection{Implementation Details}
\textbf{Data Curation}
We filter single-turn WildChat conversations exceeding 2K tokens, yielding 32K instances. We then filter long-context documents from the SlimPajama corpus into two subsets: single-document (2K–30K tokens) and multi-document (2K–20K tokens). For multi-document, we pair two same-type documents and concatenate them. We sample 100K single-document and 50K multi-document examples, totaling 150K samples.

\textbf{Training Details.} We fine-tune Mistral-7B-Instruct-v0.2 and Llama-3.1-8B-Instruct using our curated dataset. For Mistral-7B-Instruct-v0.2, we adjust the RoPE base from 1e6 to 1e7 to support longer positional embeddings\footnote{Increasing RoPE base supports longer context. More details can be seen in Appendix~\ref{sec:appendix_rope}.}. Both models are optimized using the Adam optimizer, with learning rates of 1e-6 and 5e-7 respectively. Training is conducted for 2 epochs with a batch size of 512\footnote{More details about computational budget and and infrastructure can be found in Appendix~\ref{sec:appendix_exp_details}.}.

\subsection{Baselines}
\textbf{Proprietary Long-Context Models.}
We include two proprietary long-context models \texttt{Gemini-1.5-Pro} and \texttt{GPT-4} as upper bounds due to their strong performance on long-context tasks.

\textbf{Open-Sourced Pretrained Long-Context Models.}
Additionally, we evaluate open-source pretrained language models with long-context capabilities, including GLM4-9B~\citep{glm2024chatglm}, Yi-34B~\citep{ai2024yi}, Llama3.1-70B~\citep{dubey2024llama}, Phi-3-medium~\citep{abdin2024phi3report}, and Qwen2.5~\citep{qwen2.5report}.

\textbf{Specialized Long-Context Optimized Models.}
We compare our approach to specialized long-context LLMs that extend or optimize model capabilities for long inputs. FILM~\citep{an2024make} and ChatQA-2~\citep{xu2024chatqa} fine-tunes Mistral-7B-Instruct-v0.2 and Llama-3-8B with synthetic long-context QA pairs. SEALONG~\citep{li2024sealong} applies preference optimization on Llama-3.1-8B-Instruct with extended-context QA pairs, while ProLong~\citep{gao2024prolong} continue-pretrain Llama-3-8B-Instruct to 512K context window and finetune with short-context data.

\textbf{Prior Long-Context SFT Data.}
We fine-tune Llama-3.1 on open-source long-context instruction-tuning datasets. LongAlpaca~\citep{chenlonglora} comprises 9K curated long QA pairs and 3K short QA pairs, covering tasks such as book questions and summarization. LongAlign~\citep{bai2024longalign} includes QA pairs generated by Claude 2.1 from extended documents, while LongReward~\citep{zhang2024longreward} similarly uses GLM4 to produce long-context QA pairs via a self-instruct framework.

\begin{table}[]
\caption{Main evaluation results of our models on RULER, HELMET and Longbench-Chat compared with baselines. Results on RULER and HELMET are averaged over sequence lengths ranging from 4K to 128K and 8K to 128K respectively.
}
\scalebox{0.75}{
\begingroup
\renewcommand{\arraystretch}{1.1}
\centering
\begin{tabular}{lc|ccccc|ccccccc|c}
\toprule
\multirow{2}{*}{\textbf{Models}} & \multirow{2}{*}{\textbf{Size}}               & \multicolumn{5}{c|}{\textbf{RULER}}  & \multicolumn{7}{c|}{\textbf{HELMET}}  & \multicolumn{1}{c}{\textbf{Long}} \\
            &     & \textbf{NIAH} & \textbf{VT}   & \textbf{Agg}  & \textbf{QA}   & \textbf{Avg}      & \textbf{RAG}  & \textbf{ICL}  & \textbf{Cite} & \textbf{Rank} & \textbf{QA} & \textbf{Summ} & \textbf{Avg} & \textbf{Chat}  \\
\midrule
\rowcolor{gray!20}
\multicolumn{15}{c}{\it{Proprietary Long-Context Models}}\\
\midrule
Gemini-1.5-Pro   & -    & 99.7 & 99.9 & 96.6 & 77.2 & 93.4     & 72.1 & 78.8 & 44.5 & 69.0 & 47.6 & 38.5 & 58.4 & 7.6 \\
GPT-4            & -    & 95.4 & 99.9 & 93.4 & 70.3 & 89.8     & 70.6 & 65.1 & 24.9 & 53.4 & 47.7 & 32.6 & 49.1 & 8.4 \\
\midrule
\rowcolor{gray!20}
\multicolumn{15}{c}{\it{Open-Sourced Pretrained Long-Context Models}}\\
\midrule
GLM-4-1M         & 9B   & 98.2 & 99.4 & 72.2 & 69.4 & 84.8     & 67.9 & 77.3 & 31.4 & 41.7 & 44.2 & 28.8 & 48.6 & 5.9 \\
Yi-200k          & 34B  & 95.1 & 93.6 & 74.3 & 67.1 & 82.5     & 64.1 & 78.6 & \textcolor{white}{0}4.8  & 33.4 & 25.1 & 12.2 & 36.4 & 4.0 \\
Llama-3.1        & 70B  & 96.1 & 93.2 & 83.3 & 67.8 & 85.1     & 68.6 & 77.2 & 32.9 & 52.2 & 46.0 & 33.3 & 51.7 & 6.7 \\
Phi-3-medium     & 14B  & 88.7 & 76.5 & 77.4 & 59.3 & 75.5     & 58.9 & 67.0 & 17.1 & 23.9 & 22.4 & 26.6 & 36.0 & 5.2 \\
Qwen2.5          & 7B   & 83.3 & 81.7 & 73.2 & 57.0 & 73.8     & 53.1 & 75.8 & 17.7 & 31.2 & 28.4 & 28.1 & 39.1 & 5.8 \\
\midrule            
\rowcolor{gray!20}
\multicolumn{15}{c}{\it{Specialized Long-Context Optimized Models}}\\
\midrule
FILM             & 7B   & 81.7 & 92.8 & 64.9 & 63.0 & 75.6     & 52.6 & 78.0 & 6.4  & 28.0 & 26.9 & 22.1 & 35.7 & 4.9 \\
ProLong-512k     & 8B   & 98.5 & 97.8 & 69.4 & 65.5 & 82.8     & 67.2 & 76.4 & 14.4 & 39.1 & 36.7 & 25.9 & 43.3 & 5.9 \\
ChatQA-2         & 8B   & 97.1 & \textbf{98.1} & 66.8 & 53.6 & 78.9     & 63.2 & \textbf{81.3} & \textcolor{white}{0}2.9 & 23.7 & 36.2 & 13.9 & 36.9 & 3.7 \\
SEALONG          & 8B   & 98.4 & 91.0 & 66.6 & 66.1 & 80.5     & 64.9 & 78.5 & 19.6 & \textbf{45.0} & 36.2 & 30.1 & 45.7 & 6.6 \\

\midrule
Mistral          & 7B   & 72.6 & 74.4 & 64.4 & 52.2 & 65.9     & 47.1 & 63.6 & 8.2  & 25.0 & 19.2 & 20.3 & 30.6 & 4.5 \\
+ WildLong       & 7B   & 95.2 & 95.9 & 67.0 & 64.2 & 80.6     & 62.1 & 74.6 & 12.4 & 34.3 & 34.4 & 29.2 & 41.2 & 6.3 \\
\midrule
LLaMA 3.1        & 8B   & 98.1 & 91.6 & 66.2 & 66.1 & 80.5     & 66.1 & 77.4 & 18.5 & 39.0 & 37.1 & 28.0 & 44.5 & 6.2 \\
+ WildLong       & 8B   & \textbf{98.7} & 95.7 & \textbf{74.3} & \textbf{67.9} & \textbf{84.1}     & \textbf{67.6} & 78.8 & \textbf{22.6} & 40.8 & \textbf{38.5} & \textbf{30.8} & \textbf{46.5} & \textbf{6.8} \\
\bottomrule

\end{tabular}
\endgroup}
\label{tab:overall_performance}
\end{table}

\subsection{Evaluation Benchmarks}
To comprehensively evaluate the performance of our model, we assess both long-context and short-context capabilities. For long-context tasks, we benchmark our model against established baselines, whereas for short-context tasks, we compare its performance with the base model used for fine-tuning.

For long-context tasks, we use three benchmarks designed to test a wide range of capabilities across varying input lengths:

\textbf{RULER} \citep{hsieh2024ruler}. This benchmark evaluates four synthetic task types across input lengths ranging from 4K to 128K tokens, including Needle-in-a-haystack (NIAH) retrieval, Multi-hop Tracing with Variable Tracking (VT), Aggregation (Agg), and Question Answering (QA).

\textbf{HELMET} \citep{yen2025helmet}. We evaluate our model on six tasks from HELMET: Retrieval-augmented generation (RAG), Generation with citations (Cite), Passage re-ranking (Re-rank), Long-document question answering (LongQA), Summarization (Summ), and Many-short in-context learning (ICL). The Recall task is excluded due to its overlap with the synthetic NIAH task in RULER.

\textbf{Longbench-Chat} \citep{bai2024longalign}. This benchmark tests instruction-following abilities over long contexts (10K to 100K tokens) using real-world queries. It includes 40 queries in English and 10 in Chinese. \texttt{GPT-4-128K} serves as an impartial judge to evaluate model-generated responses.

For short-context tasks, we assess general language understanding and reasoning using MMLU \citep{hendrycksmeasuring}, Winogrande \citep{sakaguchi2020winogrande}, ARC-C \citep{clark2018think}, and GSM8K \citep{cobbe2021gsm8k}, and evaluate instruction-following capabilities with IFEval\footnote{Details on evaluation settings are in Appendix \ref{sec:appendix_eval_settings}.} \citep{zhou2023instruction}.

\subsection{Results}

\textbf{Our finetuned models demonstrates strong performance over established models.} 
We significantly improve upon our baseline models, with Mistral-7B gaining +14.7 and +10.6 points on RULER and HELMET, and Llama-3.1-8B gaining +3.6 and +2.0.
Against open-source long-context models, our Llama-3.1-8B matches or exceeds larger alternatives. 
Notably on LongBench-Chat, our Llama-3.1-8B model outperforms most established models except for proprietary ones.
We also outperform specialized long-context methods. Despite using ten times more data, FILM scores lower than our Mistral-7B (e.g., 75.6 vs. 80.6 on RULER). SEALONG, based on Llama-3.1-8B-Instruct achieves lower scores, with an 8-point deficit on RULER compared with our Llama-based model. ProLong and ChatQA-2 perform well on synthetic tasks but struggle with real-world queries and complex tasks. 
These results highlight the effectiveness of our framework.

\textbf{Our method enhances performance compared to other long-context instruction-tuning data.}
We compare our method with previous long-context instruction-tuning datasets, including LongAlpaca, LongAlign and LongReward. We finetune Llama-3.1-8b-instruct with all these datasets with the same hyperparameters. As demonstrated in Table 
\ref{tab:other_long_data}, these datasets yield only slight improvements, with scores of 81.4, 81.9, and 81.2 on RULER, respectively. In contrast, our method significantly improves performance across all tasks, achieving an average score of 84.1 on RULER.
The substantial improvements in aggregation tasks, which involve integrating multiple relevant details, can likely be attributed to our dataset's focus on detail-oriented summarization and information synthesis, as illustrated in Figure~\ref{fig:task_doc_type_distribution}. This broad coverage appears to better equip models for complex long-context reasoning.
This suggests our dataset’s diversity better equips models for complex reasoning and aggregation tasks.
\begin{wraptable}{r}{0.5\textwidth}
\caption{Performance comparison of Llama-3.1-8B-instruct fine-tuned with various long-context instruction-tuning datasets.}
\vspace{0.1in}
\centering
\resizebox{0.5\textwidth}{!}{
\begin{tabular}{l|ccccc}
\toprule
\multirow{2}{*}{\textbf{Models}}    & \multicolumn{5}{c}{\textbf{RULER}}   \\
                 & \textbf{NIAH} & \textbf{VT}   & \textbf{Agg}  & \textbf{QA}   & \textbf{Avg}         \\
\midrule
LongAlpaca     & 97.9	& 95.2 & 67.0 & 65.4 & 81.4 \\
LongAlign      & 98.5	& 94.8 & 65.7 & \textbf{68.5} & 81.9 \\
LongReward     & 98.4	& 94.2 & 65.6 & 66.7 & 81.2 \\
WildLong       & \textbf{98.7} & \textbf{95.7} & \textbf{74.3} & 67.9 & \textbf{84.1} \\
\bottomrule
\end{tabular}}
\label{tab:other_long_data}
\end{wraptable}


\begin{figure}
    \centering
    \includegraphics[width=1.0\linewidth]{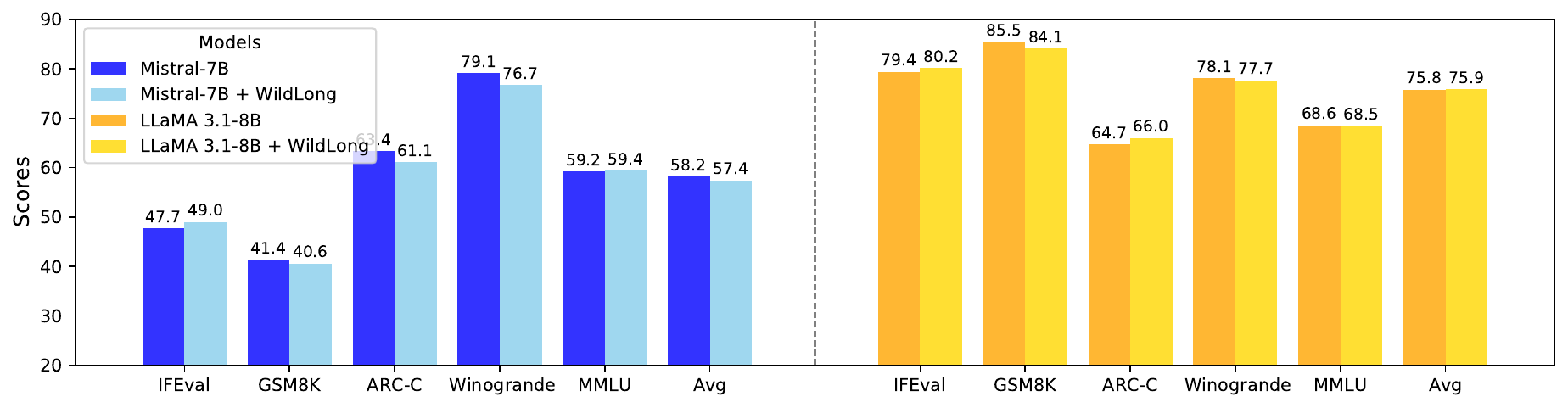}
    \caption{Comparison of short-context performances between finetuned and the baseline models. Models fine-tuned with our method preserve short-context capabilities.}
    \label{fig:short_performance}
\end{figure}

\textbf{Short context performance is preserved without mixing short-context data.} Previous works \citep{an2024make, bai2024longalign, zhang2024longreward} mix short-context instruction-tuning data into the finetuning data to mitigate degradation in short-context capabilities after long-context alignment. 
In contrast, our approach exclusively employs long-context data while effectively preserving short-context performance. Referring to Figure \ref{fig:short_performance}, we maintain an average score of 75.9 for Llama-3.1-8B, comparable to the baseline 75.8. For Mistral-7B, we observe a slight drop of less than one point, potentially due to changes in RoPE base. We analyze this further in Section \ref{mistral_analysis_rope}.
These results underscore the effectiveness of our dataset: finetuning on general, realistic long-context data significantly enhances long-context capabilities while largely preserving short-context performance without additional data mixing.


\begin{wrapfigure}{r}{0.5\textwidth}
    \includegraphics[width=0.9\linewidth]{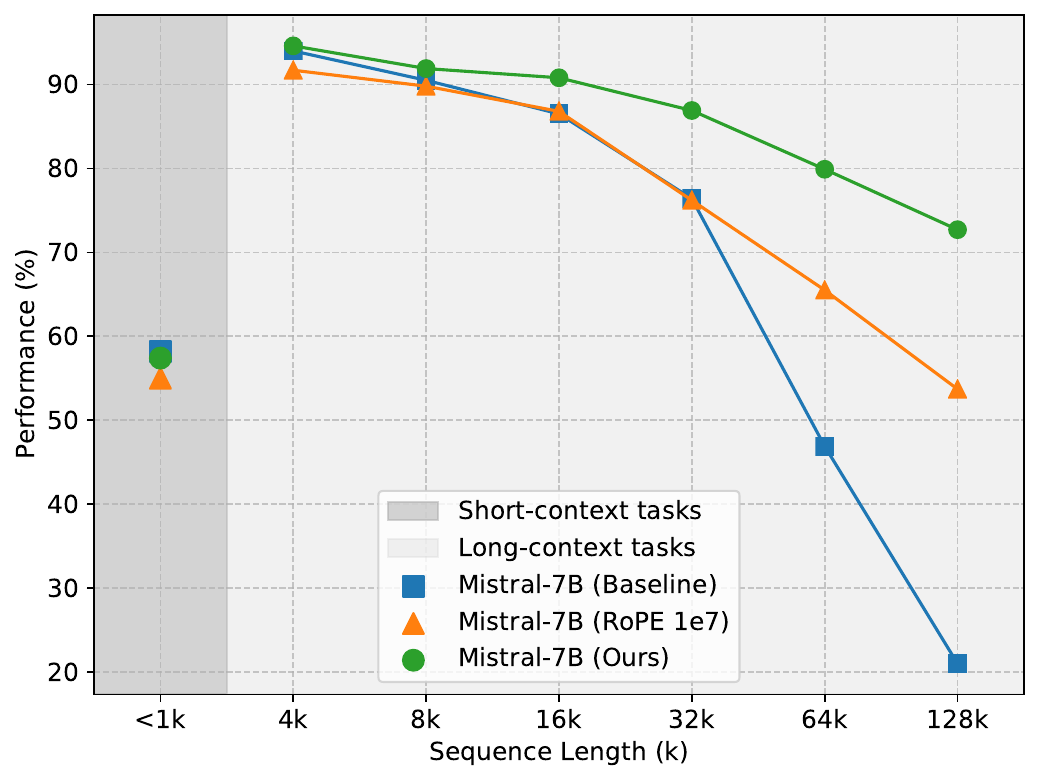}
    \caption{Short-context and long-context performance of variations of Mistral models. }
    \label{fig:performance_length}
\end{wrapfigure}

\subsection{Ablation Studies}
\label{mistral_analysis_rope}
We conduct comprehensive ablation studies to investigate the efficacy of our data synthesis framework.

\textbf{Effectiveness of graph-based modeling.}
To evaluate the effectiveness of our graph-based instruction generation approach, we compare it against two baseline methods for synthesizing long-context instruction-tuning datasets, using 20k samples for each setting. The first baseline, denoted as simple-instruct, directly extracts instructions from user-chat conversations in WildChat and pairs them with long documents sampled from SlimPajama. The second baseline, denoted as WildChat-long, finetunes directly on samples from the filtered long WildChat subset.
We finetune both Mistral-7B-Instruct-v0.2 and Llama-3.1-8b-instruct using these three datasets and evaluated them on the RULER benchmark. As shown in Table \ref{tab:ablation_graph}, our graph-based method consistently outperforms the baselines. In particular, Mistral-7B achieves a score of 78 with WildLong, outperforming WildChat-Long and Simple-Instruct by +5 and +3.8 points. We suspect the improved performance, particularly on complex tasks like aggregation and variable tracking, arises from the graph-based method’s ability to generate more diverse and challenging instructions while preserving generalizability. 

\textbf{Effectiveness of multi-document data.}
We assess the impact of multi-document data by fine-tuning both Mistral and Llama models on 20k datasets across three settings: single-document, multi-document, and a mix of both. As shown in Table \ref{tab:ablation_multi}, the results reveal varying effects depending on the model and task.
For the Mistral models, the multi-document setting significantly enhances performance on tasks requiring complex reasoning, such as variable tracking (VT) and aggregation (Agg). In contrast, single-document data proves more effective for QA tasks, which focus on extracting specific information from a single source.
For the Llama models, the effect of multi-document data is less pronounced. The multi-document setting performs slightly worse on aggregation (69.3 vs. 70.5) and QA (67.0 vs. 68.2) compared to the single-document setting. However, the mixed setting achieves the highest performance on variable tracking (93.7 vs. 93.0 for both single and multi) and matches the single-document setting in overall average performance (82.6).
These findings suggest that while single-document and multi-document data have distinct strengths, combining them provides greater diversity and balance, enabling models to perform robustly across a wide range of tasks.


\textbf{Effectiveness of WildLong under RoPE Scaling.} 
We investigate the impact of modifying the RoPE base parameter to extend the context length of the Mistral-7B model. Specifically, we compare three variants: (1) Mistral-7B (Baseline): The original model with context length 32k and RoPE base $1e6$, (2) Mistral-7B (RoPE $1e7$): Extended RoPE base of $1e7$, and (3) Mistral-7B (Ours):  RoPE base $1e7$, finetuned with our WildLong data. Performance is evaluated on short-context tasks (\textless 1k) and long-context tasks (4k-128k).

The length-wise performance is shown in Figure \ref{fig:performance_length}. Our results reveal that increasing the RoPE base parameter enables support for longer contexts, improving performance on tasks requiring extreme context lengths (e.g., 64k-128k, +18.6 and +32.7 points over the baseline Mistral-7B respectively). 
However, this adjustment comes with a significant trade-off, as short-context performance drops markedly from 58.2 to 55.0, and performance on mid-range context lengths (4k-8k) also slightly declines. 
Finetuning with WildLong mitigates these trade-offs, recovering short-context performance to 57.4 while further boosting mid- and long-context performance.
This analysis highlights that while extending RoPE theta directly allows models to process longer contexts, it introduces a clear trade-off in short-context capability. Finetuning with generalized long-context datasets, such as Wildlong, not only recovers some short-context degradation but also enhances performance across mid-range and extended contexts. These findings address a gap in prior research and emphasize the importance of finetuning strategies to balance short- and long-context performance effectively.

\begin{table}[ht]
\centering
\begin{minipage}[b]{0.45\textwidth}
\centering
\caption{Effect of graph-based modeling adopted by WildLong compared with two baseline methods.}
\resizebox{\textwidth}{!}{
\begin{tabular}{ll|ccccc}
\toprule
\multirow{2}{*}{\textbf{Model}} & \multirow{2}{*}{\textbf{Dataset}} & \multicolumn{5}{c}{\textbf{RULER}} \\
                       &                          & \textbf{NIAH} & \textbf{VT} & \textbf{Agg} & \textbf{QA} & \textbf{Avg} \\
\midrule
\multirow{3}{*}{Mistral}   & WildChat-long     & 87.7	& 84.2	 & 59.9	 & 60.3	 & 73.0 \\
                           & Simple Instruct   & 89.5	 & 90.2	 & 52.6	 & \textbf{64.3}	 & 74.2 \\ 
                           & WildLong          & \textbf{91.4}	 & \textbf{92.0}	 & \textbf{64.7}	 & 63.9	 & \textbf{78.0} \\
\midrule
\multirow{3}{*}{LLaMA}     & WildChat-long     & 98.2	& 93.5	 & 69.3	 & 67.7	 & 82.2 \\
                           & Simple Instruct   & 98.5	 & \textbf{94.1}	 & 67.6	 & \textbf{68.7}	 & 82.2 \\ 
                           & WildLong          & \textbf{98.9}	 & 93.7	 & \textbf{70.0}	 & 67.7	 & \textbf{82.6} \\
\bottomrule
\end{tabular}
\label{tab:ablation_graph}
}
\end{minipage}%
\hspace{0.5cm} 
\begin{minipage}[b]{0.45\textwidth}
\centering
\caption{Performance comparison among single-document, multi-document, and a mixture of single- and multi-document data.}
\resizebox{\textwidth}{!}{
\begin{tabular}{ll|ccccc}
\toprule
\multirow{2}{*}{\textbf{Model}} & \multirow{2}{*}{\textbf{Dataset}} & \multicolumn{5}{c}{\textbf{RULER}} \\
                       &                          & \textbf{NIAH} & \textbf{VT} & \textbf{Agg} & \textbf{QA} & \textbf{Avg} \\
\midrule
\multirow{3}{*}{Mistral}   & Single            & 91.6	& 90.9	& 63.9	& 64.2	& 77.7 \\
                           & Multi             & 92.1	 & 94.4	 & 66.9	 & 64.1	 & 79.4 \\
                           & WildLong          & 91.4	 & 92.0	 & 64.7	 & 63.9	 & 78.0 \\
\midrule               
\multirow{3}{*}{LLaMA}     & Single            & 98.6	& 93.0	& 70.5	& 68.2	& 82.6 \\
                           & Multi             & 98.8	 & 93.0	 & 69.3	 & 67.0	 & 82.0 \\
                           & WildLong          & 98.9	 & 93.7	 & 70.0	 & 67.7	 & 82.6 \\
\bottomrule
\end{tabular}
\label{tab:ablation_multi}}
\end{minipage}
\end{table}

\section{Related Work}
\textbf{Long-context Extending of LLMs.}
Several methods attempt to extend context windows with minimal training overhead. Position extrapolation approaches (\cite{chen2023extendingcontextwindowlarge, pengyarn, su2021roformer, dinglongrope, chen2024clex, liu-etal-2024-e2, zhu2024pose, wu2024skipalign, longrecipe}) adjust positional embeddings or apply rope-based techniques to accommodate longer inputs. Others manipulate attention mechanisms to scale context length (\cite{jin2024selfextend, xiao2024sink, xiao2024infllm, ding2023longnet, antraining, an2025why}), ensuring model capacity for extended sequences without complete retraining.
A separate line of work focuses on novel architectures designed for efficient long-context modeling. These include methods like Jamba (\cite{lieber2024jamba}), Unlimiformer (\cite{bertsch2024unlimiformer}), and other enhancements (\cite{wang2024augmenting, yen2024long}) to handle large inputs without quadratic complexity scaling.
Some methods rely on significant resources to equip LLMs with long-context capabilities. Llama3.1 (\cite{dubey2024llama}) conducts extensive continued pretraining on 800B tokens plus targeted fine-tuning on long-context data, and GLM (\cite{glm2024chatglm}) uses human-annotated datasets for supervised fine-tuning and DPO. While effective, these strategies can be labor-intensive or costly.
To mitigate data constraints, synthetic long-context datasets have been explored. For instance, \cite{an2024make} synthesizes QA for short context and concatenates short contexts to form long contexts, while \cite{zhao2024longskywork} synthesizes long tables to improve long-context reasoning. \cite{xu2024chatqa} uses NarrativeQA to construct long contexts by combining semantically related paragraphs, offering task-specific solutions.
Structured approaches have targeted specific long-context tasks. \cite{chen2024essential} model document correlations to curate multi-hop datasets and generate QA pairs from intra-document data. \cite{bai2024longalign} leverage Self-Instruct to create long-context instruction datasets but restrict prompts to four task types, limiting task generalization.
\cite{artificialneedles} trains the model on synthetic key-value retrieval data to improve multi-document reasoning. These methods show promise but often remain narrow in focus or require substantial manual or computational effort.
Recent approaches like LongPO \citep{chen2025longpo} and SEALONG \citep{li2024sealong} have shown that LLMs can self-improve on long-context tasks, particularly in contextual QA. LongPO extends short-context capabilities to long contexts through self-generated preference data, while SEALONG uses multiple output sampling and preference optimization to refine model responses. However, these methods focus primarily on QA tasks and do not address the broader range of challenges requiring full-context reasoning. Our approach, WildLong, is orthogonal to these methods, offering a scalable way to generate generalized data for diverse long-context tasks. 

\textbf{Scaling synthetic data creation}
Previous work on synthetic data creation for alignment has focused on leveraging human interactions with LLMs \citep{DatabricksBlog2023DollyV2, zhao2024wildchat, zheng2024lmsyschatm, pf2023openassistant}. However, manually crafting instructions is time-consuming and labor-intensive. Recent approaches have scaled instruction datasets by prompting LLMs to generate synthetic instructions, starting with a small set of human-annotated seed instructions \citep{yu2023metamath, wang2023selfinstrct, taori2023stanfordalapca, xu2024wizardlm, sun2024principle}.
Keypoints-driven strategies (\cite{li2024syntheticdataalmostscratch, tangmathscale, huang2024key}) enrich prompts with diverse topics or knowledge bases. PersonaHub \cite{ge2024personahub} introduces billions of personas to maximize coverage.
We follow this keypoints-driven philosophy but focus on long-context data synthesis, extracting meta-information from real-world conversations to generate diverse, realistic instructions closely tied to document context. By integrating document type–specific details, our framework provides a scalable route for creating high-quality long-context training data without excessive manual overhead.

\section{Conclusion}
We propose WildLong, a framework for synthesizing diverse, scalable, and realistic instruction-response datasets for long-context tasks. 
It integrates meta-information extraction to ensure realistic complexity, graph-based modeling for systematic instruction expansion, and adaptive instruction generation for enhanced contextual relevance.
Our fine-tuned models consistently outperform baselines and maintain short-context performance without mixing short-context data. Notably, our finetuned Llama-3.1-8B model surpasses most open-source long-context models on Longbench-Chat and demonstrates competitive performances with even larger models across benchmarks.
WildLong enables the synthesis of instruction-tuning data that produces robust models capable of handling diverse long-context tasks. Extending beyond synthetic QA and summarization, it bridges the gap to more complex, realistic challenges, advancing the effectiveness of long-context LLMs.
We hope WildLong provides insights into generalizing synthetic data and inspires further progress in long-context reasoning for LLMs.

\section{Limitations}
While WildLong advances synthetic data generation for long-context tasks, several limitations warrant consideration. First, although the framework mimics ``realistic'' instruction-response pairs, synthetic data may lack the nuanced complexity, ambiguity, or cultural specificity inherent in human-generated interactions. This could limit the model’s ability to handle edge cases or interpret context-dependent subtleties in real-world scenarios. Second, biases embedded in the source meta-information extracted from real user queries—such as language preferences, cultural assumptions, or domain-specific imbalances—risk propagating into the generated dataset, potentially reinforcing societal or structural inequities. Finally, while graph-based modeling captures co-occurrence relationships between entities, it may oversimplify semantic or causal dependencies, leading to superficial multi-document reasoning. 
These limitations highlight the need for hybrid data pipelines combining synthetic and human-curated examples, alongside rigorous bias audits, to enhance robustness.

\bibliography{reference}
\bibliographystyle{nat_bib}

\clearpage
\appendix
\section{Experimental Details}
\label{sec:appendix_exp_details}

\subsection{Document Classifier}
\label{sec:appendix_doc_classifier}
We trained a random forest classifier on semantic features extracted from a small language model. The classifier was trained on annotations of 20,000 long documents from SlimPajama, achieving 90\% accuracy on a held-out test set. 
Specifically, we annotated $20,000$ long documents sampled from SlimPajama using {\tt GPT-4}. The annotation prompt explicitly required the output to match one of the predefined document types, ensuring consistency with the categories defined during meta information clustering. Using these annotations, we trained a random forest classifier on semantic features extracted with StableLM-2-1.6B \citep{bellagente2024stablelm}, where the mean of the last layer's hidden states was used as the feature representation. The classifier achieved 90\% accuracy on a held-out test set, enabling efficient and accurate predictions of document types for unseen SlimPajama data.

\subsection{Instruction Generation with Paths}
\label{sec:appendix_demo}
To synthesize instructions aligned with sampled meta information paths, we prompt {\tt GPT-4} with a one-shot demonstration derived from seed paths extracted from the WildChat long conversations. 
Each seed path includes all meta information fields and a corresponding simplified instruction.
Given a sampled path $\hat{P}$, we identify the most similar seed path $P^*$ based on the of their nodes. The similarity between paths is computed as $\text{intersection\_sim}(\hat{P}, P^*) = | \hat{P} \cap P^* |.$ The selected example path and its instruction are included in the prompt to guide {\tt GPT-4} in generating a new instruction given a new path. This ensures the generated instruction aligns with the sampled meta information criteria, while benefiting from the contextual relevance provided by the seed example.
{\tt GPT-4} synthesizes a natural language instructions adhering to the sampled path's constraints with the prompt shwon in Table \ref{tab:prompt_path_to_instruct}.

\subsection{Evaluation Settings}
\label{sec:appendix_eval_settings}
For short-context evaluation, we utilize the \texttt{lm-evaluaton-harness} framework \cite{eval-harness} and following the evaluation settings in \citep{open-llm-leaderboard}: 25-shots for ARC-C, and 5-shots for MMLU, Winogrande and GSM8K. We use 0-shot for IFEval. We report the \texttt{acc\_norm} metric for ARC-C, the \texttt{acc} metric for MMLU, Winogrande and GSM8K. We average the metrics \texttt{prompt\_level\_strict\_acc}, \texttt{inst\_level\_strict\_acc}, \texttt{prompt\_level\_loose\_acc}, and \texttt{inst\_level\_loose\_acc} for IFEval. 

For long-context evaluations, we evaluate our models and all baselines following the settings in the original benchmarks. Table \ref{tab:appendix_model_info}  presents the sources of evaluation results for the models across three benchmarks.

\begin{table}[h]
\caption{Evaluation source for each model on three benchmarks. 
    \cmark\ indicates that the evaluation was conducted by ourselves, while \ostar\ indicates that results were sourced from the original benchmark. 
}
\centering
\scalebox{0.8}{
\begin{tabular}{l|ccc}
\toprule
\textbf{Models} & \textbf{RULER} & \textbf{HELMET} & \textbf{Longbench-Chat} \\
\midrule
\rowcolor{gray!20}
\multicolumn{4}{c}{\it{Proprietary Long-Context Models}}\\
\midrule
Gemini-1.5-Pro   & \ostar  & \ostar  & \cmark  \\
GPT-4            & \ostar  & \ostar  & \ostar  \\
\midrule
\rowcolor{gray!20}
\multicolumn{4}{c}{\it{Open-Sourced Pretrained Long-Context Models}}\\
\midrule
GLM-4-1M         & \ostar  & \cmark  & \cmark  \\
Yi-200k          & \ostar  & \ostar  & \cmark  \\
Llama-3.1-70B    & \ostar  & \ostar  & \cmark  \\
Phi-3-medium     & \ostar  & \ostar  & \cmark  \\
Qwen2.5          & \cmark  & \cmark  & \cmark  \\
Mistral-7B       & \cmark  & \cmark  & \cmark  \\
Llama-3.1-8B     & \cmark  & \cmark  & \cmark  \\
\midrule            
\rowcolor{gray!20}
\multicolumn{4}{c}{\it{Specialized Long-Context Optimized Models}}\\
\midrule
FILM             & \cmark  & \cmark  & \cmark  \\
ProLong-512k     & \cmark  & \cmark  & \cmark  \\
ChatQA-2         & \cmark  & \cmark  & \cmark  \\
SEALONG          & \cmark  & \cmark  & \cmark  \\
\bottomrule
\end{tabular}
}
\label{tab:appendix_model_info}
\end{table}

\subsection{Technical Details}
We employ several open-source libraries and tools for model training. Specifically, we use PyTorch \citep{paszke2019pytorch} and the Hugging Face Transformers library \citep{wolf2019huggingface} for implementing and fine-tuning the model. To enhance computational efficiency, we integrate FlashAttention 2 \citep{dao2023flashattention2} for optimized attention computation. The fine-tuning process is conducted on eight AMD Radeon Instinct MI300 GPUs, each equipped with 192GB of memory. Training on 150K synthetic data samples requires approximately 480 GPU hours.

\section{Discussion on RoPE Base}
\label{sec:appendix_rope}
Recent studies demonstrate that adjusting the base value in Rotary Position Embedding (RoPE) significantly enhances language models' ability to handle long-context sequences \citep{ntkawarerope}. 
The scalability of RoPE for long-context has been systematically demonstrated through base parameter adjustments \citep{liuscaling}.
By increasing the base parameter (e.g., from $10^4$ to $10^6$), the wavelength of positional encoding grows exponentially as \( \lambda_i \propto \text{base}^{2i/d} \), where \(d\) is the embedding dimension and \(i\) indexes frequency bands. This prolongs the non-repeating positional patterns across distant tokens, effectively mitigating encoding collisions that impair long-range dependency modeling. Practical implementations like Code Llama \citep{grattafiori2023code} and ChatGLM \citep{glm2024chatglm} have adopted base scaling to extend context windows to 16k+ tokens while preserving local positional sensitivity. Our experiments align with these findings, showing that larger base values improve coherence in tasks requiring cross-sentence reasoning.
\section{Prompts}
\label{sec:appendix_prompt}

The prompts used by the WildLong framework can be seen from Table~\ref{tab:prompt_extract_meta}, Table~\ref{tab:prompt_path_to_instruct}, Table~\ref{tab:prompt_instruct_response}, and Table~\ref{tab:prompt_single_to_multi}.

\begin{table}[h!]\centering
\begin{minipage}{\textwidth}
\centering
\begin{tcolorbox} 
    \centering
   
      \small
    \begin{tabular}{p{0.95\textwidth}}
    Below is a conversation between a user and an AI Language Model, likely involving a long document.\\
    \\ \textbf{Conversation}
         \\ {\tt \{conversation\}} \\ 
         \\ \textbf{Your Tasks}
         \\ Based on the conversation above, try to finish the following tasks.
         \\ - Determine whether the query of the user involves a long document (or any form of long text). 
         \\ - If the conversation involves a long document, analyse the conversation and provide the following information using concise phrases. 
         \\ \hspace{0.4em} - Document Type: Specify the format or category of the document, such as a research paper, technical report, fictional story, instruction manual, etc. Ideally, extract one document type. However, if you believe there are multiple types, limit the number to two.
         \\ \hspace{0.4em} - Tasks or Requests: Identify 1 to 3 the specific tasks the user wants the chatbot to perform given the long context. This may include summarizing key points, integrating multiple pieces of information, continuing the dialogue or story, providing an analysis, or any other specific task relevant to the long text.
         - Purpose of Query: Define the objective behind the user's query, such as educational purposes, decision-making, research, entertainment, etc. List 1 to 3 items.
         \\ \hspace{0.4em} - User Intention: Determine the underlying goal or reason behind the user's request, such as completing an assignment, preparing for a debate, gaining a general understanding, etc. List 1 to 3 items.
         \\ \hspace{0.4em} - User Profile: Describe the possible characteristics and background of the user in 1 to 5 phrases.
         \\ \hspace{0.4em} - User's Language Style: Identify the language style of the user. List 1 to 3 items.
         \\ \hspace{0.4em} - Context: Describe the situational background influencing the query, such as working on a group project, preparing for an exam, etc. List 1 to 3 context items.
         \\ \hspace{0.4em} - Knowledge/Commonsense Involved for User: Identify the prior knowledge or commonsense the user is expected to have. List 1 to 5 items.
         \\ \hspace{0.4em} - Knowledge/Commonsense Involved for Chatbot: Identify the prior knowledge or commonsense the chatbot is expected to have to address the query. List 1 to 5 items.
         \\ \hspace{0.4em} - Long Context Capability Involved: Determine the comprehension and information processing skills required to address the user's request, such as long document comprehension, key information retrieval, handling multiple perspectives, etc. List 1 to 3 items.
         \\ \hspace{0.4em} - Output Format: Identify the desired format of the response. List 1 to 3 items.
         \\ \hspace{0.4em} - Sentiment: Determine the expected emotional tone or attitude in the response. List 1 to 3 items.
         \\ \hspace{0.4em} - Constraint of the Request: Identify the limitations or additional requirements that the user has for the chatbot's response. List 0 to 3 constraints, if any.
         \\ \hspace{0.4em} - Simplified Instruction by User: Provide a simplified version of the user's request, removing any context or background information.

    \\ \textbf{Output Format}
        \\ Document Type:
        \\ 1. doc type 1 ...
        \\ 2. doc type 2 ...
        \\
        \\ Task or Request:
        \\ 1. request type 1 ...
        \\ 2. request type 2 ...
        \\ ...

    \\ \textbf{Additional Requirements for Output}
        \\ - Analyze the entire conversation to produce your answers, taking into account both the user's and the chatbot's contributions. Do not limit your analysis to just one side.
        \\ - If the user query does not involve a long document (or any form of long text), output only "No long document involved".
        \\ - For each output field, output commonly used phrases or short sentences in academic or industry if applicable.
        \\ - If you cannot extract anything for a particular field, output "NA" for that field.
    \end{tabular}
\end{tcolorbox}
\caption{The prompt to extract meta information with GPT-4.}
    \label{tab:prompt_extract_meta}
\end{minipage}
\end{table}

\begin{table}[h!]\centering
\begin{minipage}{\textwidth}
\centering
\begin{tcolorbox} 
    \centering
   
      \small
    \begin{tabular}{p{0.95\textwidth}}
    You are tasked with generating 3 realistic user queries or instructions for a chatbot about a long document. The user is interacting with a long {\tt \{doc{\_}type\}}, but you do not have access to the exact content of the document. Your task is to create reasonable user queries or instructions that meet specific meta information criteria. There are 12 meta information categories that define the characteristics of a user query or instruction. You will be provided with 6 key meta information fields that must be incorporated into each of your generated queries or instructions. For the remaining 6 categories, you have the flexibility to explore different possibilities to create varied and diverse queries or instructions. You will be given an example meta information criteria and a corresponding sample query or instruction to help you understand the context and how to apply the meta information.\\
    \\
    \textbf{Additional requirements}\\
    - Incorporate All Key Fields: Aim to integrate all 6 key meta information fields into each query or instruction you create. If a field is particularly challenging to include, substitute it with a reasonable alternative.\\
    - Ensure Coherence and Creativity: Your generated queries or instructions should be coherent, natural, and flow smoothly. They should not appear as a direct combination of the meta information fields, instead aiming for a realistic scenario that a user in the given context might actually encounter.\\
    - Creative Interpretation: The meta information criteria represent high-level characteristics of a user's query or instruction. You can interpret and apply them creatively to generate a range of realistic and diverse outputs.\\
    - Output Format: Present your generated queries or instructions in bullet points, formatted as follows:\\
    1. {{query 1}}\\
    2. {{query 2}}\\
    3. {{query 3}}\\
    \\
    \textbf{Definitions of the 12 meta information categories}\\
    - Tasks or Requests: tasks the user wants the chatbot to perform given the long context.\\ 
    - Purpose of Query: the objective behind the user's query, such as educational purposes, decision-making, research, entertainment, etc.\\
    - User Intention: the underlying goal or reason behind the user's request, such as completing an assignment, preparing for a debate, gaining a general understanding, etc.\\
    - User Profile: the possible characteristics and background of the user.\\
    - User's Language Style: the language style of the user.\\
    - Context: the situational background influencing the query, such as working on a group project, preparing for an exam, etc.\\
    - Knowledge/Commonsense Involved for User: the prior knowledge or commonsense the user is expected to have.\\
    - Knowledge/Commonsense Involved for Chatbot: the prior knowledge or commonsense the chatbot is expected to have to address the query.\\
    - Long Context Capability Involved: the comprehension and information processing skills required to address the user's request, such as long document comprehension, key information retrieval, handling multiple perspectives, etc.\\
    - Output Format: the desired format of the response.\\
    - Sentiment: the expected emotional tone or attitude in the response.\\
    - Constraint of the Request: the limitations or additional requirements that the user has for the chatbot's response.\\
    \\
    \textbf{Example meta information criteria}\\
    {\tt \{example{\_}meta{\_}info\}}\\
    \\
    \textbf{Example query/instruction}\\
    {\tt \{example{\_}instruction\}}\\
    \\
    \textbf{Your task}\\
    Generate a new query or instruction that aligns with the given meta information criteria:\\
    {\tt \{path{\_}meta{\_}info\}}\\
    \end{tabular}
\end{tcolorbox}
\caption{The prompt to generate instruction given a sampled meta-information path.}
    \label{tab:prompt_path_to_instruct}
\end{minipage}
\end{table}

\begin{table}[h!]\centering
\begin{minipage}{\textwidth}
\centering
\begin{tcolorbox} 
    \centering
   
      \small
    \begin{tabular}{p{0.95\textwidth}}
    \textbf{Long Document:}\\
    {\tt \{long{\_}doc\}}\\
    \\
    \textbf{Example Query/Instruction:}\\
    {\tt \{example{\_}instruct\}}\\
    \\
    \textbf{Your Task:}\\
    You have been provided with a long document above, along with an example query or instruction that was formulated for another similar long document.\\
    
    Your task is to create a new query or instruction that can be addressed using the information contained within the long document provided.\\
    
    The new query or instruction should be inspired by the structure and intent of the given example but is not a direct copy. You should adapt the query or instruction to fit the context of the long document while still addressing a similar type of task.\\
    
    Once you have formulated the query or instruction, provide a response based on the content of the long document.\\
    \\
    Please format your output as follows:\\
    Query/Instruction: \{\{query{\_}or{\_}instruction\}\}\\
    Response: \{\{response\}\}
    \end{tabular}
\end{tcolorbox}
\caption{The prompt to generate instruction-response pairs.}
    \label{tab:prompt_instruct_response}
\end{minipage}
\end{table}

\begin{table}[h!]\centering
\begin{minipage}{\textwidth}
\centering
\begin{tcolorbox} 
    \centering
   
      \small
    \begin{tabular}{p{0.95\textwidth}}
    The following are tasks or requests made by users when querying a chatbot about a single document. Modify the tasks or requests as if the user is querying multiple documents. Ensure that the modifications reflect a realistic need to handle information across multiple sources, incorporating cognitive operations usually applied to multiple documents.\\
    The document type is {\tt \{doc{\_}type\}}. Avoid simply adding phrases like "across multiple documents." Instead, adapt each task to reflect a more complex interaction with multiple sources, focusing on the cognitive operation that makes sense in the multi-document context.\\
    \\
    \textbf{Cognitive operations}
    - Comparison: identifying similarities, differences, or evaluating multiple documents\\
    - Synthesis: integrating information from multiple sources to create a new, cohesive understanding\\
    - Aggregation: collecting and presenting information from multiple sources without integrating or interpreting\\
    - Verification and Validation: cross-referencing and fact-checking across documents\\
    - Consensus Analysis: identifying agreement across documents\\
    - Divergence Analysis: recognizing conflicting or differing points of view\\
    - Problem Solving: formulating solutions based on multiple documents\\
    - Decision Making: formulating decisions based on multiple documents\\
    - Exploration: discovery across multiple sources without a predefined goal\\
    - Trend and Pattern Identification: detecting larger patterns or trends from multiple documents\\
    - Hypothesis Generation: forming new hypotheses through integrated data\\
    - Creative Synthesis: fostering novel ideas or concepts from the documents\\
    \\
    \textbf{Original tasks or requests}
    {\tt \{original{\_}tasks{\_}or{\_}requests\}}\\
    \\
    \textbf{Output format}\\
    1. { \{original{\_}tasks{\_}or{\_}requests\}}: { \{modified{\_}tasks{\_}or{\_}requests\}}\\
    2. { \{original{\_}tasks{\_}or{\_}requests\}}: { \{modified{\_}tasks{\_}or{\_}requests\}}\\
    ...\\
    \end{tabular}
\end{tcolorbox}
\caption{The prompt to convert single-document tasks to multi-document tasks. }
    \label{tab:prompt_single_to_multi}
\end{minipage}
\end{table}

\end{document}